\documentclass[lettersize,journal]{IEEEtran}
\usepackage{amsmath,amsfonts}
\usepackage{amssymb}
\usepackage{algorithmic}
\usepackage{algorithm}
\usepackage{array}
\usepackage[caption=false,font=normalsize,labelfont=sf,textfont=sf]{subfig}
\usepackage{textcomp}
\usepackage{stfloats}
\usepackage{url}
\usepackage{verbatim}
\usepackage{multirow}
\usepackage{subfig}
\usepackage{graphicx}
\usepackage{caption,subcaption}
\usepackage{booktabs}
\usepackage{cite}
\usepackage{xcolor}
\usepackage{subcaption}
\usepackage{subfloat}
\begin{document}

\title{LoopExpose: An Unsupervised Framework for Arbitrary-Length Exposure Correction}

\author{Ao Li, Chen Chen, Zhenyu Wang, Tao Huang, Fangfang Wu, Weisheng Dong,~\IEEEmembership{Member, IEEE}

\thanks{Ao Li and Weisheng Dong are with the School of Artificial Intelligence, Xidian University, Xi’an 710071, China (e-mail: ali\_0607@stu.xidian.edu.cn; wsdong\@mail.xidian.edu.cn). 

Zhenyu Wang and Tao Huang are with the Hangzhou Institue of Technology, Xidian University, Hangzhou 311231, China.

Fangfang Wu is with the School of Computer Science and Technology, Xidian University, Xi'an 710071, China.

Chen Chen is with the School of Information Science and Engineering, Dalian University of Technology, Dalian 116000, China.
}}

\markboth{Journal of \LaTeX\ Class Files,~Vol.~14, No.~8, August~2021}%
{Shell \MakeLowercase{\textit{et al.}}: A Sample Article Using IEEEtran.cls for IEEE Journals}

\IEEEpubid{0000--0000/00\$00.00~\copyright~2021 IEEE}

\maketitle

\begin{abstract}
Exposure correction is essential for enhancing image quality under challenging lighting conditions. While supervised learning has achieved significant progress in this area, it relies heavily on large-scale labeled datasets, which are difficult to obtain in practical scenarios. To address this limitation, we propose a pseudo label-based unsupervised method called LoopExpose for arbitrary-length exposure correction. A nested loop optimization strategy is proposed to address the exposure correction problem, where the correction model and pseudo-supervised information are jointly optimized in a two-level framework. Specifically, the upper-level trains a correction model using pseudo-labels generated through multi-exposure fusion at the lower level. A feedback mechanism is introduced where corrected images are fed back into the fusion process to refine the pseudo-labels, creating a self-reinforcing learning loop. Considering the dominant role of luminance calibration in exposure correction, a Luminance Ranking Loss is introduced to leverage the relative luminance ordering across the input sequence as a self-supervised constraint. Extensive experiments on different benchmark datasets demonstrate that LoopExpose achieves superior exposure correction and fusion performance, outperforming existing state-of-the-art unsupervised methods. Code is available at \url{https://github.com/FALALAS/LoopExpose}.
\end{abstract}


\begin{figure*}[t]
	\centering
	\includegraphics[width = \linewidth]{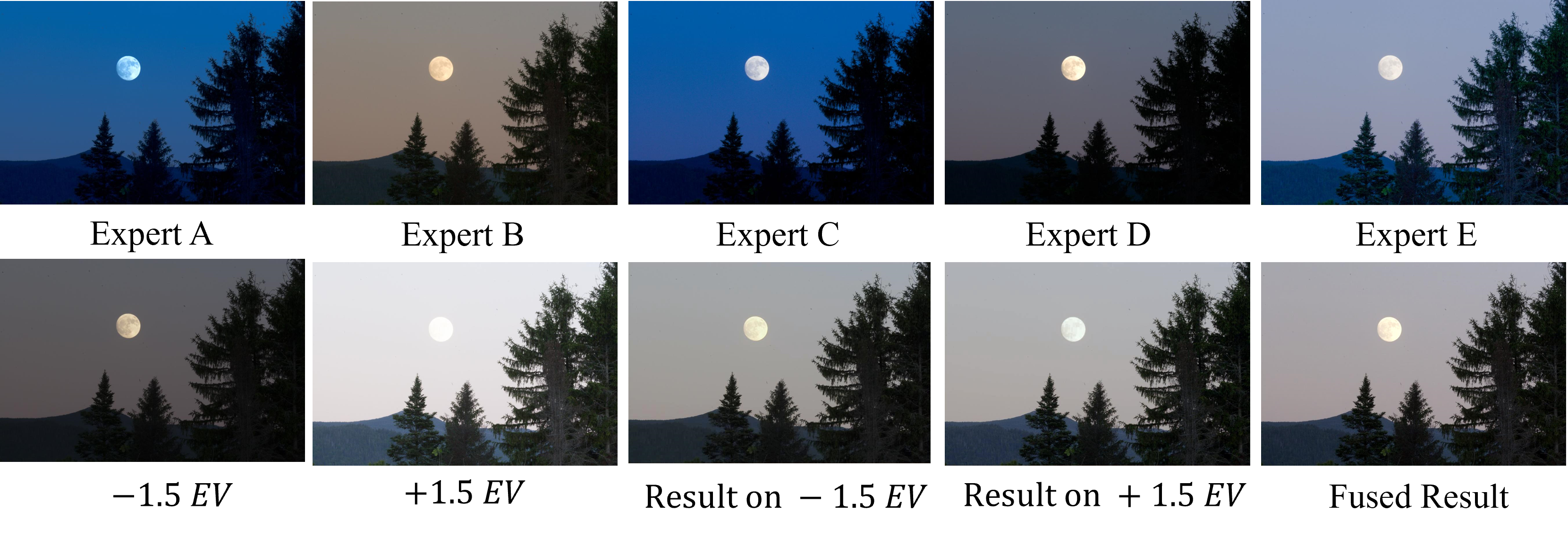}
	\caption{Illustration of stylistic diversity among expert-retouched reference images in the MSEC dataset \cite{msec}. The figure also shows the error-exposed inputs with exposure values (EV) of $-1.5$ and $+1.5$, together with the correction result produced by our LoopExpose. While human experts exhibit varying tone and contrast preferences, our method faithfully enhances the original scene, demonstrating strong robustness against subjective editing biases.} 
	
	\label{experts}
\end{figure*}

\section{Introduction}
Exposure plays a vital role in photography, directly affecting image clarity and overall visual quality. While modern cameras attempt automatic exposure, challenging lighting often causes errors that degrade images and harm downstream vision tasks.

Existing deep learning-based exposure correction methods \cite{msec, cotf, lact, mmht, fecnet, da} have achieved impressive results but heavily rely on large-scale paired datasets. These datasets are costly to create and, as shown in Figure~\ref{experts}, suffer from annotator-specific stylistic biases that limit model generalization. A recent unsupervised method \cite{uec} attempts to reduce manual annotation by transforming exposure information from reference images into error-exposed images. As exposure correction inherently requires both underexposed and overexposed images for effective training, treating each image independently overlooks the intrinsic relationships among varying exposure levels. Consequently, the potential consistency and complementary information within exposure sequences remain underexplored.

Studies~\cite{noise2noise, noise2void, neighbor2neighbor} in self-supervised learning for low-level vision tasks have demonstrated that supervision can be derived directly from corrupted data, \emph{e.g.}, Noise2Noise~\cite{noise2noise} achieves denoising by predicting the statistical mean of noisy distributions. Inspired by this, we indicate that the exposure error can also be regarded as a degradation like noise and explore the analogous strategies for it. Different from random noises, exposure deviations exhibit a more complex distribution, making it ineffective to offset deviations by simply getting means of them. Fortunately, the prior multi-exposure fusion (MEF) algorithms \cite{mertens, gf} which recover a high-quality image by aggregating differently exposed ones, implicitly suggesting that the well-exposed image hides in the error exposure sequence.

Based on the aforementioned insights, we propose an unsupervised exposure correction framework that conducts the exposure correction learning with pseudo-supervisors from sequence fusion, refrained from manual editing bias and enabling more robust and scalable exposure correction. 
To overcome the quality limits of static pseudo-labels, we introduce a nested loop optimization strategy that jointly refines the exposure correction model and pseudo-labels in a two-level framework.
Specifically, the upper-level trains the exposure correction model on current pseudo-labels while the lower-level processes regenerates better labels by fusing the newly-corrected images. This design enables both the model and supervision signals to evolve together. As a result, our unsupervised and progressive learning strategy achieves high-fidelity correction without handcrafted ground truth.

In the exposure correction branch, lumainance and color are respectively corrected with the luminance-aware module and the 3D LUT module to reduce mutual interference and then intergrated with the attention-based fusion. Considering the central role of luminance calibration in exposure correction, we introduce a Luminance Ranking Loss that exploits the natural luminance ordering within multi-exposure sequences as a self-supervised constraint. This loss captures exposure relationships across the sequence, encouraging the network to learn luminance-aware representations.

\IEEEpubidadjcol

In real-world applications, large-scale training datasets can be efficiently constructed from batches of RAW files by applying fixed exposure offsets, without requiring manual annotations. Common RAW processing tools, such as the Adobe Camera Raw SDK~\cite{adobe}, allow efficient generation of a multi-exposure sequence from a single RAW input. We evaluate our framework on two widely used exposure correction benchmarks, MSEC \cite{msec} and Radiometry512 \cite{uec}. To enhance their usability, we reorganize both datasets into more flexible and standardized formats suitable for diverse training and evaluation protocols. Extensive experiments across multiple benchmarks demonstrate that our method outperforms state-of-the-art approaches both qualitatively and quantitatively. Overall, our main contributions are as follows:

\begin{itemize}
	\item We propose LoopExpose, a novel unsupervised exposure correction framework based on a self-reinforcing learning loop, significantly improving correction performance without the requirement for manually annotated data.
	\item We introduce a sequence-based self-supervised Luminance Ranking Loss, leveraging the relative illuminance information within exposure sequences to effectively guide exposure correction.
	\item We present a new unsupervised baseline framework for arbitrary-length exposure correction and reorganize existing public datasets into versatile formats to construct effective training sets and facilitate future research.
\end{itemize}

\section{Related Works}

\subsection{Single-Exposure Correction}

Single-exposure correction (SEC) aims to restore visual information in a single incorrectly exposed image by enhancing contrast and dynamic range. Traditional methods, such as histogram equalization \cite{clahe, adaptivehe}, curve mapping \cite{autoec}, and Retinex models \cite{wvm, naturalnessretinex}, rely on handcrafted priors and often struggle with complex real-world scenarios.

Early learning-based methods primarily address low-light image enhancement, focusing on underexposure correction via Retinex-based decomposition \cite{anisotropicRetinex, retina, noiseretinex,prdiffretinex,diffdarkretinex}, curve estimation \cite{zerodce, zerodce++}, or direct image-to-image mapping \cite{ctnet, exposurediff, gamma}. However, these approaches typically generalize poorly to overexposed cases.

A paradigm shift occurrs with the seminal work of Afifi et al. \cite{msec}, who propose a unified SEC framework that corrects both under- and overexposed regions. Subsequent research follows several directions. One is exposure-invariant feature learning, where methods like CMEC \cite{CMEC}, ENC \cite{enc}, ECLNet \cite{ecl}, and ERL \cite{erl} learn representations robust to varying exposure levels. Another direction explores multi-domain processing strategies. For example, FECNet \cite{fecnet} leverages frequency domain analysis by separating and processing amplitude and phase components; LACT \cite{lact} and CoTF \cite{cotf} enhance exposure correction by applying learnable transformations in color spaces, and \cite{da} introduces reparameterization with multiple convolution kernels to capture complementary representations. Additionally, hierarchical architectures have been explored, such as the multi-scale design in MSEC and the transformer-based structure in MMHT \cite{mmht}, which captures both global and local dependencies.

Recently, UEC \cite{uec} introduces the first unsupervised SEC method by borrowing the idea of style transfer \cite{cyclegan}. UEC designs a self-supervised strategy in which differently exposed images mutually supervise, further extracting exposure features from an arbitrary well-exposed reference image and applying them to incorrectly exposed images for correction. 
In contrast, we propose a brand-new unsupervised framework for exposure correction, built upon a nested optimization strategy that jointly updates the correction model and its pseudo-supervisory signals. This design empowers LoopExpose to significantly surpass prior methods, establishing a new SOTA benchmark in unsupervised exposure correction.

\subsection{Multi-Exposure Fusion}

Multi-Exposure Fusion (MEF) synthesizes high dynamic range images from multiple images with varying exposure values. Traditional models like Mertens \cite{mertens} fuse images based on hand-crafted features and remain competitive due to their simplicity and robustness. Recently, deep learning-based MEF methods \cite{deepfuse,mefgan, mefnet, meflut, transmef, ifcnn} have demonstrated superior performance in challenging multi-exposure fusion scenarios. 

Our framework can be viewed as a two-stage fusion algorithm. To ensure training stability in the nested optimization loop, we employ the Mertens fusion algorithm. Notably, when enhanced by our framework, the Mertens algorithm achieves results that surpass even SOTA deep learning-based methods.

\begin{figure*}[htbp]
	\subfloat[Overall architecture of the proposed LoopExpose framework, where the lower-level MEF model generate pseudo-labels and the upper-level SEC model enhances the input sequence in a nested loop.]{
		\centering
		\includegraphics[width=0.6335\textwidth]{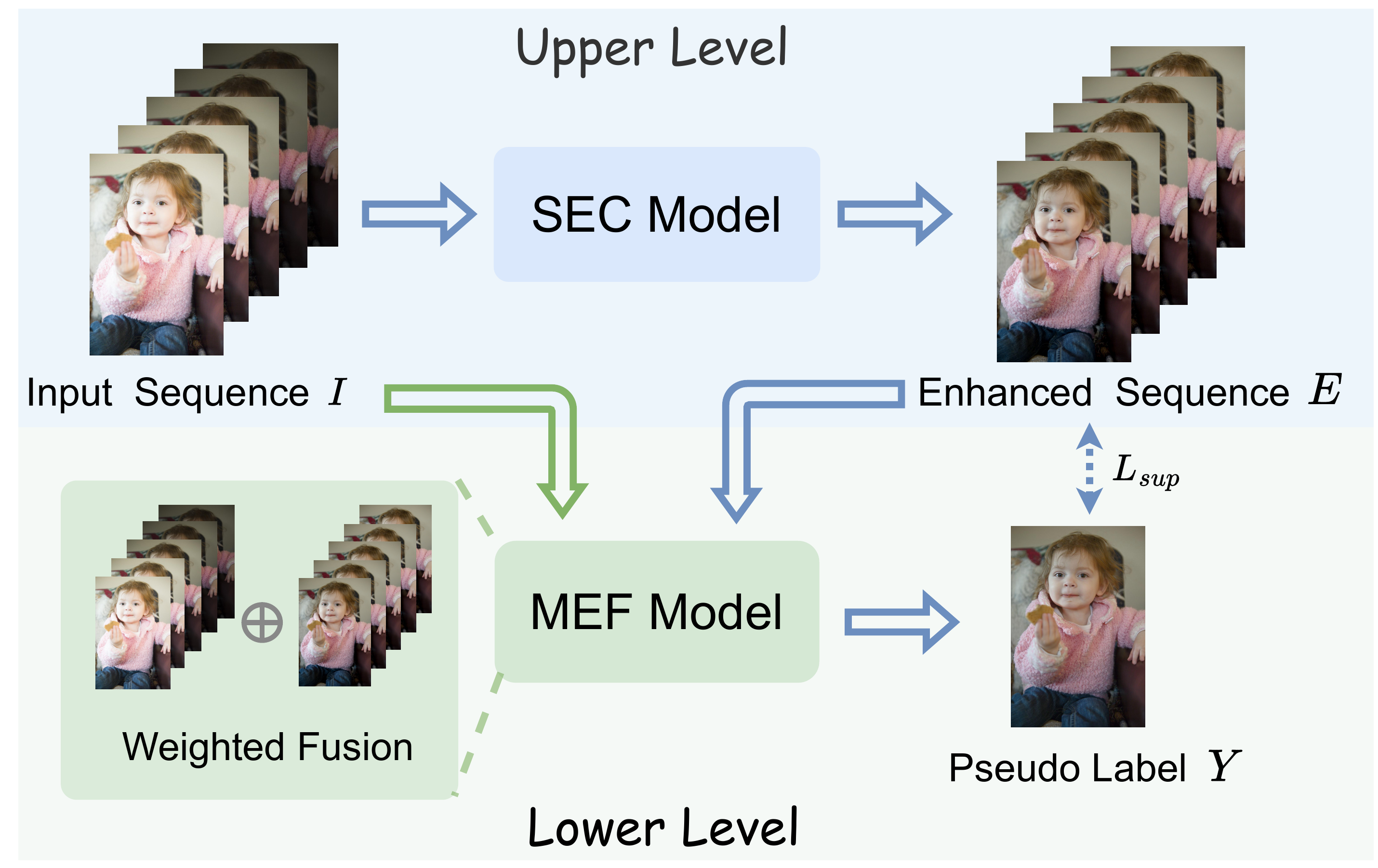}
		\label{fig:frame}
}
\hspace{0.05cm}
    \subfloat[Architecture of our SEC model. It comprises a luminance-aware network and an adaptive 3D LUT module.]{
		\centering
		\includegraphics[width=0.3565\textwidth]{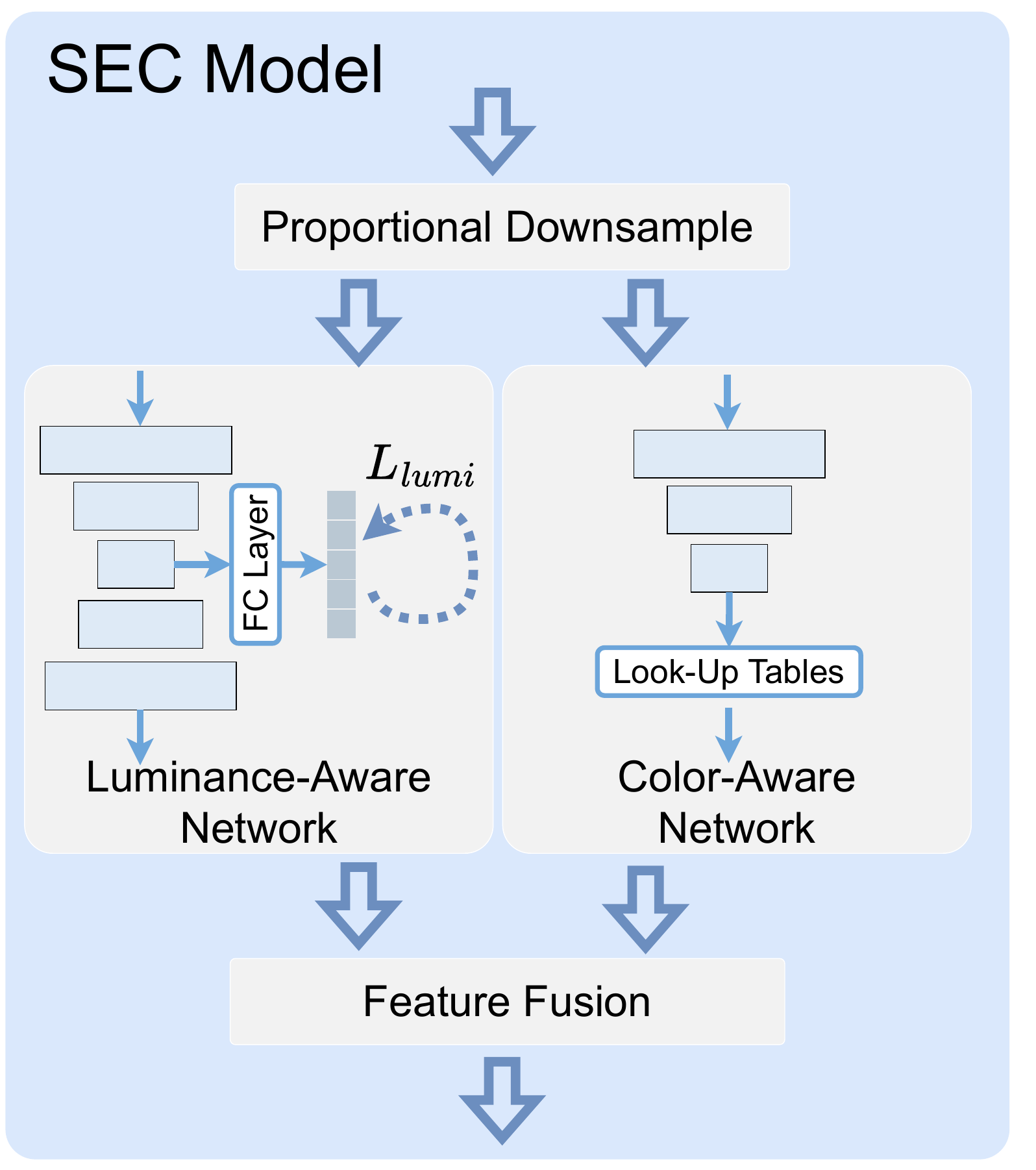}
		\label{fig:sec}
}
    \caption{Illustration of the proposed LoopExpose framework. (a) shows the training pipeline based on our nested optimization loop. (b) details the internal components of the SEC model.}
	\label{fig:framework}
\end{figure*}

\subsection{SEC meets MEF}

The dual illumination estimation method \cite{dual} is the first to introduce the concept of MEF for SEC. By constructing pseudo multi-exposure sequences through image inversion, this method breaks away from the traditional SEC paradigm, influencing the architectural design of subsequent deep learning-based approaches.
Building on this idea, LCDP \cite{lcdp} employs an encoder-decoder network to generate overexposure- and underexposure-enhanced images, which are then fused to produce the final correction. Similarly, CSEC \cite{csec} extends dual illumination estimation by generating darken and brighten features and fusing them into a pseudo-normal representation to refine exposure and mitigate color shifts.
Moving beyond the dual-illumination framework, FCNet \cite{fcnet} combines a MEF network with a SEC network to incorporate real multi-exposure sequence information into exposure correction, further improving robustness.

While previous methods integrate MEF into SEC pipelines through architectural design or feature fusion \cite{dual, lcdp, csec, fcnet}, our approach unifies them under a nested optimization loop. In our framework, the fusion module generates pseudo-labels that guide the training of the exposure correction model, whose updated outputs are in turn fed back to refine the fusion results. This iterative interaction forms a self-reinforcing correction–fusion process, enabling unsupervised learning without manual supervision and jointly improving both tasks.

\section{Methodology}

\subsection{Theoretical Background}

Recent self-supervised methods for low-level vision tasks, such as denoising and restoration, have demonstrated that meaningful supervision can be derived directly from corrupted data without relying on clean ground truth \cite{noise2noise, noise2void, neighbor2neighbor, przero}. Inspired by this, we treat the exposure error as a degradation and explore a similar learning paradigm for it. However, unlike random noise, exposure variations are significantly more complex and require more sophisticated strategies to approximate the underlying scene radiance.

In classical image denoising tasks, the observed image $I^{\text{obs}}$ is typically modeled as a clean signal $E^{\text{real}}$ corrupted by noise $ \varepsilon$:
\begin{equation} 
	I^{\text{obs}} = E^{\text{real}} + \varepsilon,
\end{equation}
A common assumption is that $ \varepsilon$ is additive white Gaussian noise, \emph{i.e.}, $\varepsilon \sim \mathcal{N}(0, \sigma^2)$ \cite{dncnn}. Under this assumption, Noise2Noise \cite{noise2noise} demonstrates that a model can effectively learn to approximate the ground truth signal by estimating the mean of the noise distribution.

In contrast to additive noise, exposure errors represent a structured, non-linear degradation. This can be described using the exposure-response model of digital imaging systems \cite{hdrmodel}, which primarily links exposure variations to exposure time. Although simplified, studies \cite{debevec, modelingResponse} show that key exposure distortions can be effectively modeled through exposure time $\Delta t$ and camera response function (CRF) $f(\cdot)$. Given multiple images with different exposure times $\Delta t_i$, the observed images $I^{\text{obs}}_i$ can be formulated as:
\begin{equation}
	I^{\text{obs}}_i = f(E^{\text{real}} \cdot \Delta t_i), \quad i = 1,2,\dots,N,
	\label{hdrmodel}
\end{equation}
where $E^{\text{real}}$ denotes the scene radiance, which remains consistent across different exposures.

Unlike Gaussian noise, exposure errors are neither zero-mean nor uniform, with their effects on pixels being highly non-linear due to the interplay of exposure time and camera response function. However, just as noisy datasets have an inherent ground truth, multi-exposure fusion algorithms \cite{mertens, gf} show that a latent ground truth can be approximated from multi-exposure sequences. These algorithms compute a content-adaptive, weighted combination of multiple exposures to approximate the latent radiance $E^{\text{real}}$, serving as a non-trivial mean in the exposure domain. Motivated by these insights, we propose a self-supervised framework based on a nested optimization loop that learns a data-driven mean to recover the underlying signal for arbitrary-length exposure correction.

\subsection{Framework Overview}

\subsubsection{Nested Loop Optimization}  
As shown in Fig.~\ref{fig:frame}, we propose using the MEF (\emph{e.g.}, Mertens \cite{mertens}) images as pseudo-labels to supervise the exposure correction model training. Since the supervisor is absolutely derived from the exposure sequence, it prevents the biases from manual retouching and ensures the model focuses solely on exposure correction without altering color characteristics. However, the pseudo-labels fused based on the initial low-quality sequences can provide limited guidance, finally restricting the correction performance.

To overcome this limitation, we formulate the learning process as a nested loop optimization strategy with two key phases, including Warm-Up and Joint Optimization. This strategy operates within a two-level framework designed to progressively refine both the correction model and its supervision. Specifically, the upper-level improves the exposure correction model, while the lower-level refines pseudo-labels by multi-exposure fusion. This process creates a self-reinforcing loop, where the lower level generates supervision for the upper level, and the improved model in turn provides higher-quality inputs back to the lower level for the next pseudo-label update.

The design of our framework intentionally deviates from conventional bilevel optimization methods \cite{traditionalBL,2021bilevel}. In these approaches, both levels are typically optimized simultaneously, which, in unsupervised settings like ours, can lead to incorrect learning directions and unstable training \cite{stabilityBL}, as illustrated in Figure~\ref{fig:mefnet}. To avoid such collapse, we restrict the lower-level of our framework to serve as a response unit. 

The upper-level improves the exposure correction model by minimizing the exposure correction loss:
\begin{equation}
	\min_{\theta} L (\theta, I, Y, E),
\end{equation}
where $\theta$ denotes the parameters of the exposure correction model, and $I$, $Y$, and $E$ represent the input images, pseudo-labels, and enhanced images, respectively. The function $L$ is the overall loss used to guide the training process.

The lower-level refines pseudo-labels $Y$ progressively:
\begin{equation}
	Y^{(t+1)} = 
	\begin{cases}
		\mathcal{M}(I), & \text{Warm-Up} \\
		\mathcal{M}(I, E^{(t)}), & \text{Joint Optimization}
	\end{cases},
\end{equation}
where $\mathcal{M}$ denotes the Mertens fusion algorithm \cite{mertens}. By employing a non-trainable rule-based fusion, we limit the flexibility of lower-level optimization, ensuring that the stability of the entire training process.

The learning phases are summarized as follows:
\begin{itemize}
	\item \textbf{Warm-Up}: Initializes the correction model to prevent distortion, using only the original exposure sequence for fusion. The learning rate decreases gradually for stable parameter updates.
	\item \textbf{Joint Optimization}: The core phase, where the correction model and fusion process interact iteratively. The learning rate is reset and maintained constant for stable joint training.
\end{itemize}

\begin{figure}[t]
	\centering
	\includegraphics[width = \columnwidth]{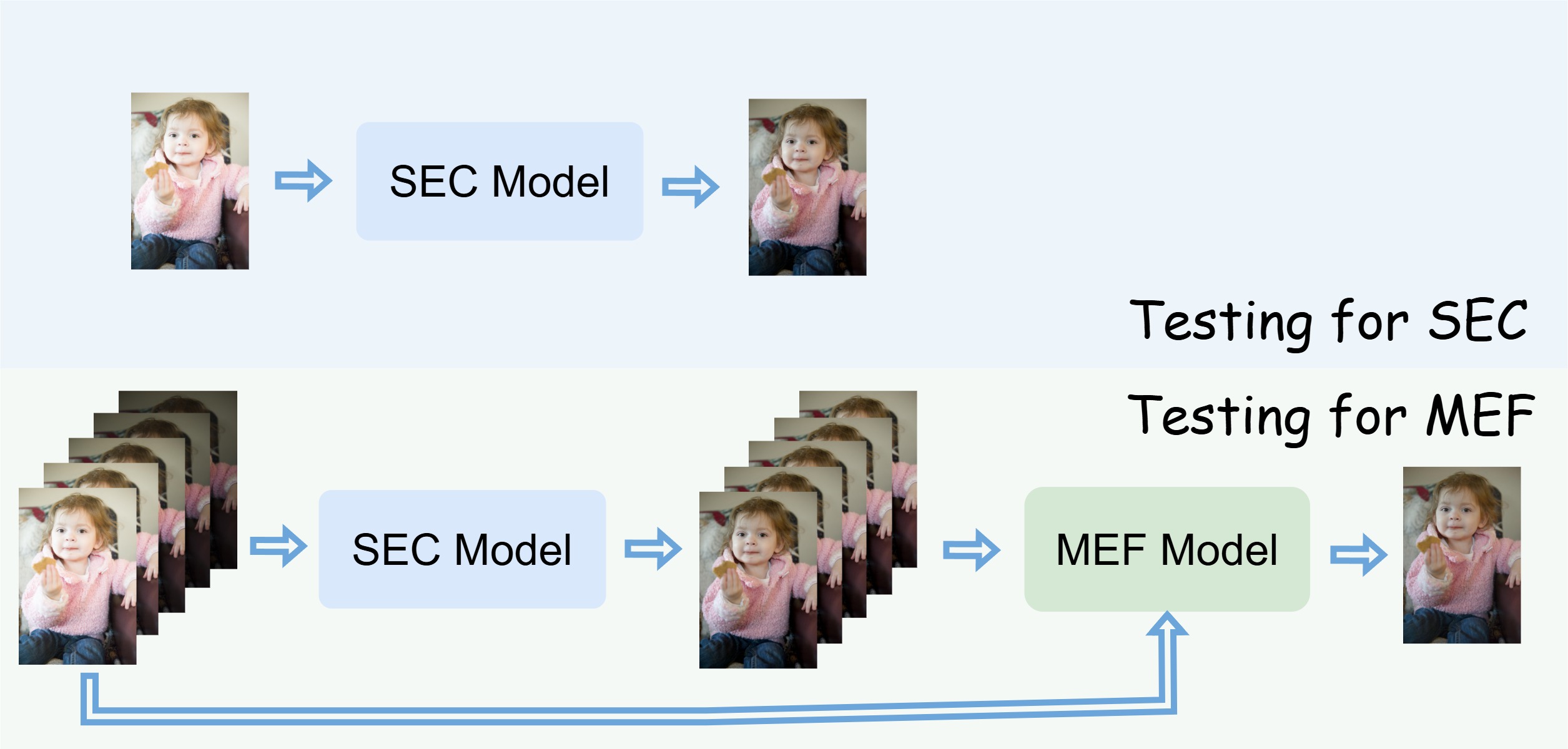}
	\caption{Testing pipeline of LoopExpose. The trained exposure correction model processes arbitrary-length exposure sequences to generate enhanced outputs for fusion or final output.} 
	\label{test}
\end{figure}

\subsubsection{Arbitrary-Length Exposure Correction}
Another key advantage of our framework is its natural ability to process exposure sequences of arbitrary lengths. 
As shown in Figure \ref{test}, an arbitrary-length exposure sequence is processed by the trained correction model to produce enhanced outputs during inference. These corrected images can be regarded as the final results when the input sequence length is one or further fused using MEF for longer sequence.
This design eliminates the requirement for fixed-length input or explicit reference images. To the best of our knowledge, LoopExpose is the first unsupervised method that can be applied to both MEF and SEC tasks. 

\subsection{Exposure Correction and Fusion Model}

Following previous design paradigms \cite{lact, cotf, prhdr}, we construct a dual-branch network. To minimize mutual interference caused by aliasing, luminance and color features are learned separately.

To guide exposure correction globally, we design a Luminance-Aware Network that extracts multi-scale luminance features using a lightweight encoder-decoder architecture.  
The network also outputs a global luminance descriptor $F^L$, summarizing the overall exposure status of the input and used in our proposed self-supervised Luminance Ranking Loss. While the Luminance-Aware Network provides global luminance guidance, local and fine-grained exposure corrections are handled by Look-Up Tables (LUTs). Following the adaptive fusion strategy in \cite{3dlut}, multiple basic 3D LUTs are dynamically combined based on the input content, enabling adaptive generation of suitable transformations for diverse exposure errors. After obtaining luminance and color features, we follow LACT \cite{lact} and employ a revised attention mechanism enabling efficient fusion of the two features and generating the final corrected images.

\begin{figure}[!t]
	\centering
	\includegraphics[width=1\linewidth]{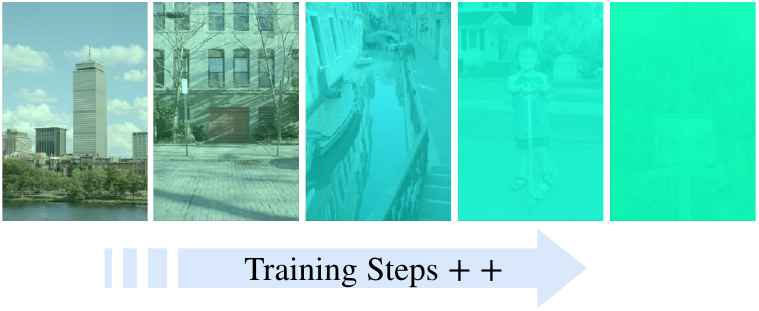}
	\caption{Illustration of error accumulation when a learnable MEF network \cite{mefnet} is used to generate pseudo-labels.}
	\label{fig:mefnet}
\end{figure}

To generate pseudo-labels within our nested optimization framework, we adopt the Mertens fusion algorithm \cite{mertens}, whose rule-based, non-trainable nature ensures stable supervision during training. Due to the nature of the loss function used in unsupervised MEF methods \cite{meflut,mefnet}, which enforces similarity between each input and the final fused image, this objective inevitably leads to error accumulation when applied in our iterative framework. Our experiments further confirm this issue. As shown in Figure \ref{fig:mefnet}, using MEFNet caused our correction model to collapse and converge to a static value.

\subsection{Loss Functions}

Our framework combines self-supervised constraints with pseudo-supervised guidance using the following composite loss:
\begin{equation}
	L_{total} = L_{lumi} + L_{sup},
\end{equation}
where $L_{lumi}$ is the self-supervised loss and $L_{sup}$ is the supervised loss based on pseudo-labels.

To encourage the learning of luminance features, we introduce a self-supervised Luminance Ranking Loss $L_{lumi}$ that leverages the inherent ordering of exposure sequences. As demonstrated in \cite{pairwisecomparison}, even when direct estimations of certain quantities are unavailable, their relative ordering or approximate ratios can often be reliably inferred. In practice, even if the original luminance order of the sequence is unknown, it can be easily approximated by ranking the input images according to their average pixel intensity. The proposed Luminance Ranking Loss enforces a monotonic constraint on luminance feature values, ensuring that darker images produce lower feature responses than brighter ones. This design effectively guides the luminance-aware network in capturing and encoding luminance-specific information in a self-supervised manner.

In our SEC model, the luminance-aware network incorporates global average pooling and fully connected layers in the encoder stage, compressing the spatial dimensions into a $N \times 1$ global descriptor $F^L$. $L_{lumi}$ is defined as follows:
\begin{equation}
	L_{lumi} = w_{lumi} \sum_{i=1}^{N-1} \sum_{j=i+1}^{N} \max\left(0,\, F^L_i + \text{margin} - F^L_j\right).
\end{equation}
The image sequence is arranged in order from dark to bright. Particularly, $w_{lumi}$ is set to 1.

The supervised loss $L_{sup}$ guides the SEC model by minimizing the discrepancy between the enhanced image $E$ and the pseudo-label $Y$ obtained via fusion. Inspired by prior methods~\cite{cotf, lact}, it is a composite loss combining L1 loss $L_1$, perceptual loss $L_p$, and SSIM loss $L_{ssim}$ as:
\begin{equation}
	L_{sup} = L_1 + w_p L_p + w_{ssim} L_{ssim},
\end{equation}
with weights $w_p = 0.1$ and $w_{ssim} = 0.05$.

\section{Experiments}
\subsection{Experimental Settings}

\subsubsection{Datasets}

We evaluate our method using two exposure correction datasets, MSEC \cite{msec} and Radiometry512 \cite{uec}, both derived from the MIT-Adobe FiveK Dataset \cite{fivek}. The MSEC \cite{msec} dataset contains sequences with relative exposure values (EVs) from -1.5 to +1.5. To better support sequence-based training, we reorganize MSEC into a sequential structure by grouping images of the same scene under different exposures into folders, forming the SeqMSEC variant.

In contrast, Radiometriy512 \cite{uec} dataset renders expert-retouched images covering a wider exposure range from -3 to +3. As this dataset lacks official splits, we partition it into training, validation, and testing sets following strategy of MSEC. To prevent GT data leakage, we remove the image of 0 EV.
Afterwards, we reorganize it in two formats: SeqRadio, a sequential version (similar to SeqMSEC) and PlainRadio, a version that mimics the original flat structure of MSEC, facilitating broader research applications.  

Importantly, in real-world applications, similar to how these datasets are constructed, large-scale training data can be easily built from batches of RAW files by applying fixed exposure offsets during export. This process requires neither manual annotations nor camera-specific calibration, making it highly practical and scalable for deployment. Common RAW development tools, such as Adobe Camera Raw SDK~\cite{adobe}, support the efficient generation of aligned multi-exposure sequences from a single RAW input. 

\subsubsection{Implementation Details}
The proposed exposure correction model is implemented in PyTorch and trained on an NVIDIA RTX 4090 GPU. As to the multi-exposure fusion model, we directly use the OpenCV \cite{opencv} implementation of Mertens algorithm. More details can be found in our code. 

\begin{table*}[!t]
	\centering
    \caption{Quantitative comparison with SOTA exposure correction methods on the MSEC \cite{msec} dataset. T, S, and U denote traditional methods, supervised methods, and unsupervised methods, respectively. \textcolor{red}{\textbf{Red}} indicates the overall best performance among all methods, while \textcolor{blue}{\textbf{Blue}} marks the best result among unsupervised approaches. All results are obtained using the released model weights by these methods.}
    \label{compareAtMSEC}
	\resizebox{\textwidth}{!}{
	\begin{tabular}{l c *{12}{r}}  
		\toprule[1.5pt]
		\multirow{2}{*}{Method} & \multirow{2}{*}{Type} 
		& \multicolumn{2}{c}{Expert A} 
		& \multicolumn{2}{c}{Expert B} 
		& \multicolumn{2}{c}{Expert C} 
		& \multicolumn{2}{c}{Expert D} 
		& \multicolumn{2}{c}{Expert E} 
		& \multicolumn{2}{c}{Avg} \\
		\cmidrule(lr){3-4} 
		\cmidrule(lr){5-6}
		\cmidrule(lr){7-8} 
		\cmidrule(lr){9-10} 
		\cmidrule(lr){11-12} 
		\cmidrule(lr){13-14}
		& 
		& PSNR & SSIM 
		& PSNR & SSIM
		& PSNR & SSIM
		& PSNR & SSIM
		& PSNR & SSIM
		& PSNR & SSIM \\
		\midrule
		
		WVM (CVPR16) & T  
		& 14.537 & 0.721 & 15.859 & 0.788 & 15.154 & 0.751
		& 15.923 & 0.779 & 16.540 & 0.807 & 15.603 & 0.769 \\
		
		Dual (CGF19) & T  
		& 15.911 & 0.731 & 17.144 & 0.787 & 16.647 & 0.758
		& 16.785 & 0.773 & 17.134 & 0.796 & 16.724 & 0.769 \\
		\midrule
		MSEC (CVPR21) & S  
		& 19.158 & 0.746 & 20.096 & 0.734 & 20.205 & 0.769 
		& 18.975 & 0.719 & 18.983 & 0.727 & 19.483 & 0.739 \\
		
		LCDPNet (ECCV22) & S  
		& 20.574 & 0.809 & 21.804 & 0.865 & 22.295 & 0.855
		& 20.108 & 0.824 & 19.281 & 0.822 & 20.812 & 0.835 \\
		
		MMHT (MM23) & S  
		& 20.662 & 0.816 & 22.377 & 0.869 & 23.049 & 0.865
		& 20.416 & 0.831 & 20.520 & 0.833 & 21.405 & 0.843 \\
		
		LACT (ICCV23) & S 
		& 20.950 & 0.821 & 22.751 & 0.858 & \textcolor{red}{\textbf{23.531}} & 0.869
		& 20.808 & 0.838 & 20.605 & 0.841 & 21.728 & 0.846 \\
		
		CoTF (CVPR24) & S  
		& \textcolor{red}{\textbf{21.167}} & \textcolor{red}{\textbf{0.827}} & \textcolor{red}{\textbf{22.923}} & 0.865 & 23.455 & \textcolor{red}{\textbf{0.872}}
		& \textcolor{red}{\textbf{20.846}} & \textcolor{red}{\textbf{0.841}} & \textcolor{red}{\textbf{20.697}} & \textcolor{red}{\textbf{0.843}} & \textcolor{red}{\textbf{21.818}} & \textcolor{red}{\textbf{0.850}} \\
		
		CSEC (CVPR24) & S  
		& 15.349 & 0.682 & 16.842 & 0.771 & 16.083 & 0.714
		& 16.020 & 0.730 & 16.377 & 0.745 & 16.134 & 0.729 \\
		\midrule
		ZeroDCE++ (TPAMI21) & U 
		& 11.643 & 0.536 & 12.555 & 0.539 & 12.058 & 0.544 & 12.964 & 0.548 & 13.769 & 0.580 & 12.597 & 0.549\\
		
		RUAS (CVPR21) & U 
		& 10.166 & 0.391 & 10.522 & 0.440 & 9.356 & 0.411 & 11.013 & 0.441 & 11.574 & 0.466 & 10.526 & 0.430\\
		
		SCI (CVPR22) & U 
		& 12.806 & 0.655 & 13.735 & 0.724 & 13.314 & 0.690
		& 14.199 & 0.715 & 14.831 & 0.743 & 13.777 & 0.705 \\
		
		PairLIE (CVPR23) & U 
		& 9.425 & 0.576 & 10.068 & 0.631 & 9.731 & 0.610
		& 10.693 & 0.639 & 11.554 & 0.673 & 10.294 & 0.626 \\
		
		UEC (ECCV24) & U 
		& 18.904 & 0.789 & 19.991 & 0.852 & 19.049 & 0.804
		& 17.943 & 0.795 & 17.187 & 0.793 & 18.615 & 0.807 \\
		
		LoopExpose (Ours) & U 
		& \textcolor{blue}{\textbf{19.597}} & \textcolor{blue}{\textbf{0.795}} & \textcolor{blue}{\textbf{22.249}} & \textcolor{red}{\textbf{0.874}} & \textcolor{blue}{\textbf{21.007}} & \textcolor{blue}{\textbf{0.826}}
		& \textcolor{blue}{\textbf{20.174}} & \textcolor{blue}{\textbf{0.830}} & \textcolor{blue}{\textbf{19.896}} & \textcolor{blue}{\textbf{0.839}} & \textcolor{blue}{\textbf{20.585}} & \textcolor{blue}{\textbf{0.833}} \\
		\bottomrule[1.5pt]
	\end{tabular}}
\end{table*}

\begin{table}[t]
	\centering
    	\caption{Quantitative comparison with SOTA exposure correction methods on the Radiometry512 \cite{uec} dataset. \textcolor{red}{\textbf{Red}} indicates the overall best performance among all methods, while \textcolor{blue}{\textbf{Blue}} marks the best result among unsupervised approaches. LACT \cite{lact} and CoTF \cite{cotf} are retrained on our PlainRadio dataset.}
        \label{compareAtRadio}
	\begin{tabular}{l c r r}  
		\toprule[1.5pt]
		Method & Type & PSNR & SSIM \\  
		\midrule
		WVM (CVPR16) & T  & 15.357 & 0.738 \\
		Dual (CGF19) & T  & 17.367 & 0.756 \\
		\midrule
		LACT (ICCV23) & S & 23.790 & 0.851 \\
		CoTF (CVPR24) & S & \textcolor{red}{\textbf{23.997}} & \textcolor{red}{\textbf{0.861}} \\
		\midrule
		ZeroDCE++ (TPAMI21)    & U & 12.729 & 0.386 \\
		RUAS (CVPR21)          & U & 11.086 & 0.539 \\
		SCI (CVPR22)           & U & 12.647 & 0.611 \\
		PairLIE (CVPR23)       & U & 13.800 & 0.659 \\
		UEC (ECCV24)           & U & 18.885 & 0.779 \\
		LoopExpose (Ours)      & U & \textcolor{blue}{\textbf{21.529}} & \textcolor{blue}{\textbf{0.821}} \\
		\bottomrule[1.5pt]
	\end{tabular}
\end{table}

\subsection{Comparisons with SOTA Methods}

Although our framework is primarily designed for SEC, it can be seamlessly extended to handle MEF as well. We evaluate its performance on both tasks and benchmark our method against the latest state-of-the-art techniques.

\subsubsection{Single-Exposure Correction Results}

The papers participating in the comparison include traditional algorithms (WVM \cite{wvm}, Dual \cite{dual}), supervised learning-based models (MSEC \cite{msec}, LCDPNet \cite{lcdp}, MMHT \cite{mmht}, LACT \cite{lact}, CoTF \cite{cotf}, CSEC \cite{csec}), and several unsupervised approaches (ZeroDCE++ \cite{zerodce++}, RUAS \cite{ruas}, SCI \cite{sci}, PairLIE \cite{pairlie}, UEC \cite{uec}). Notably, among the unsupervised methods, UEC is uniquely tailored for exposure correction, whereas the others were originally intended for low-light enhancement and thus are not optimized for exposure-variable scenes.

Despite not relying on manually curated ground truth data, LoopExpose demonstrates competitive performance. As summarized in Table~\ref{compareAtMSEC}, our unsupervised framework consistently outperforms all existing unsupervised baselines, including the previously best method, UEC \cite{uec}. Compared with the supervised baseline MSEC \cite{msec}, LoopExpose delivers superior results. Moreover, our model closely approaches the performance of the strongest supervised method, CoTF \cite{cotf}.

\begin{table}[!t]
	\centering
    \caption{Quantitative comparison with SOTA MEF methods on MSEC \cite{msec} and Radiometry512 \cite{uec} Datasets. MEFNet \cite{mefnet} and MEFLUT \cite{meflut} are retrained on our reorganized SeqMSEC and SeqRadio datasets for fair comparison.}
    \label{mef}
	\begin{tabular}{l *{4}{r}}  
		\toprule[1.5pt]
		\multirow{2}{*}{Method} 
		& \multicolumn{2}{c}{MSEC} 
		& \multicolumn{2}{c}{Radiometry512} \\
		\cmidrule(lr){2-3} \cmidrule(lr){4-5}
		& PSNR & SSIM & PSNR & SSIM \\
		\midrule
		Mertens (CGF09)      & 19.405 & 0.825 & 21.662 & 0.842 \\
		MEFNet (TIP20)       & 18.292 & 0.808 & 20.198 & 0.832  \\
		MEFLUT (ICCV23)      & 20.588 & 0.829 & 17.570 & 0.792   \\
		LoopExpose (Ours)    & \textbf{21.321} & \textbf{0.847} & \textbf{23.222} & \textbf{0.889}  \\
		\bottomrule[1.5pt]
	\end{tabular}
\end{table}

\begin{table}[t]
	\centering
	\caption{Quantitative comparison with arbitrary-Length exposure correction method FCNet \cite{fcnet} on the MSEC \cite{msec} dataset.}
	\label{tab:sec-mef}
	\begin{tabular}{ccccc}
		\toprule[1.5pt]
		\multirow{2}{*}{Method} 
		& \multicolumn{2}{c}{SEC} 
		& \multicolumn{2}{c}{MEF} \\
		\cmidrule(lr){2-3} \cmidrule(lr){4-5}
		& PSNR & SSIM & PSNR & SSIM \\
		\midrule
		FCNet (Supervised)      & 19.67 & 0.799 & 20.81 & 0.847 \\
		LoopExpose (Unsupervised) & \textbf{21.01} & \textbf{0.826} & \textbf{21.32} & \textbf{0.847} \\
		\bottomrule[1.5pt]
	\end{tabular}
\end{table}

As to the Radiometry512 dataset \cite{uec}, for fairness, supervised methods such as LACT \cite{lact} and CoTF \cite{cotf} are retrained on our plainRadio. The results in Table~\ref{compareAtRadio} confirm the strong performance of LoopExpose. Our method substantially outperforms previous unsupervised techniques and achieves performance close to the supervised state-of-the-art model CoTF \cite{cotf}. Other unsupervised models, including UEC \cite{uec}, exhibit clear limitations under diverse exposure scenarios.

\begin{figure*}[t]
	\centering
	\includegraphics[width = \linewidth]{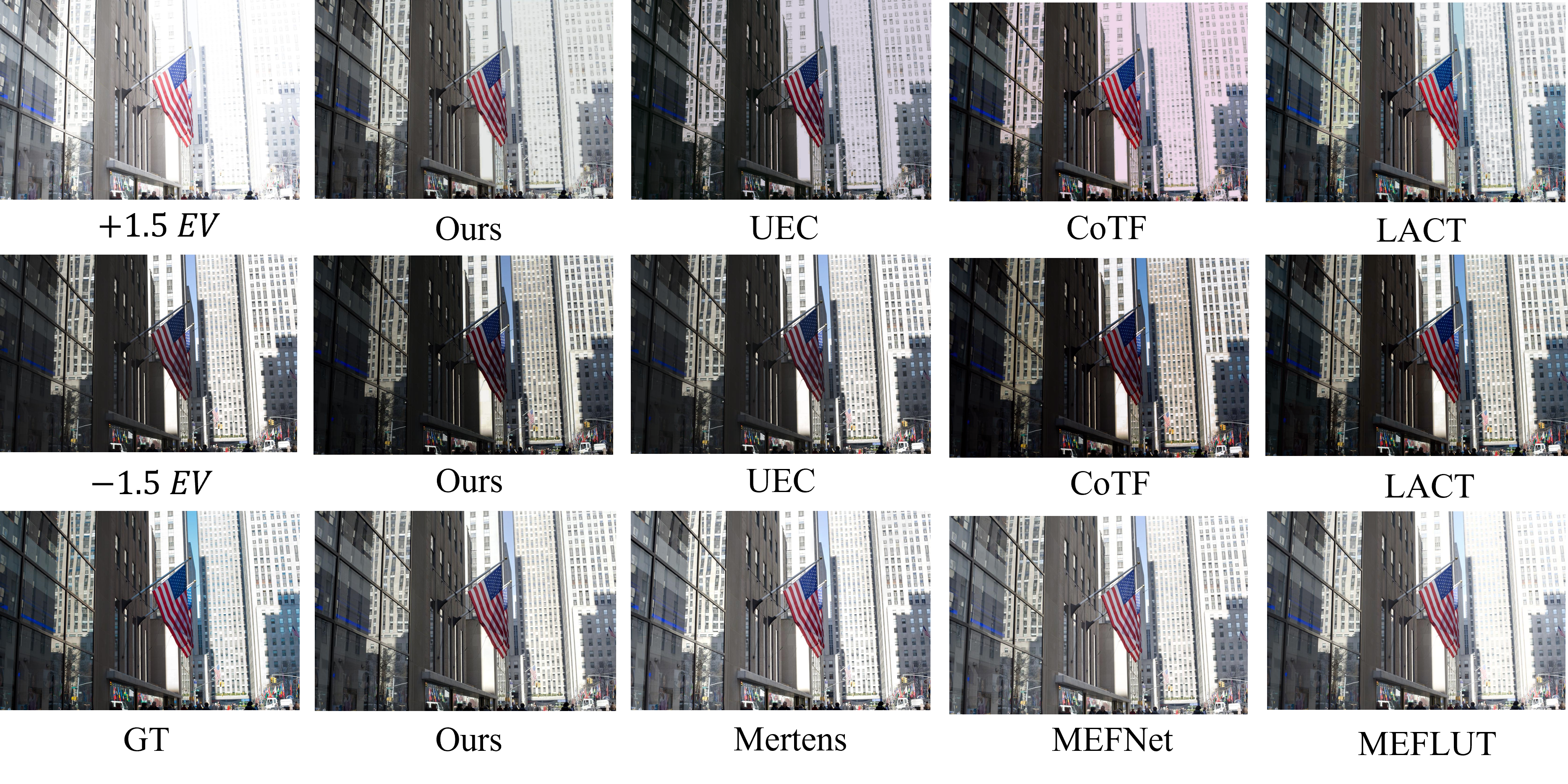}
	\caption{Visual comparison of exposure correction and fusion results on the MSEC dataset \cite{msec}.
		The first row displays the original $+1.5$ EV image and its corresponding corrected results by different methods, while the second row shows the $-1.5$ EV image and its corrected versions. The third row presents the ground truth (GT) and the fused results generated by different MEF algorithms.} 
	\label{viscom}
\end{figure*}

\begin{figure}[t]
	\centering
	\includegraphics[width = \linewidth]{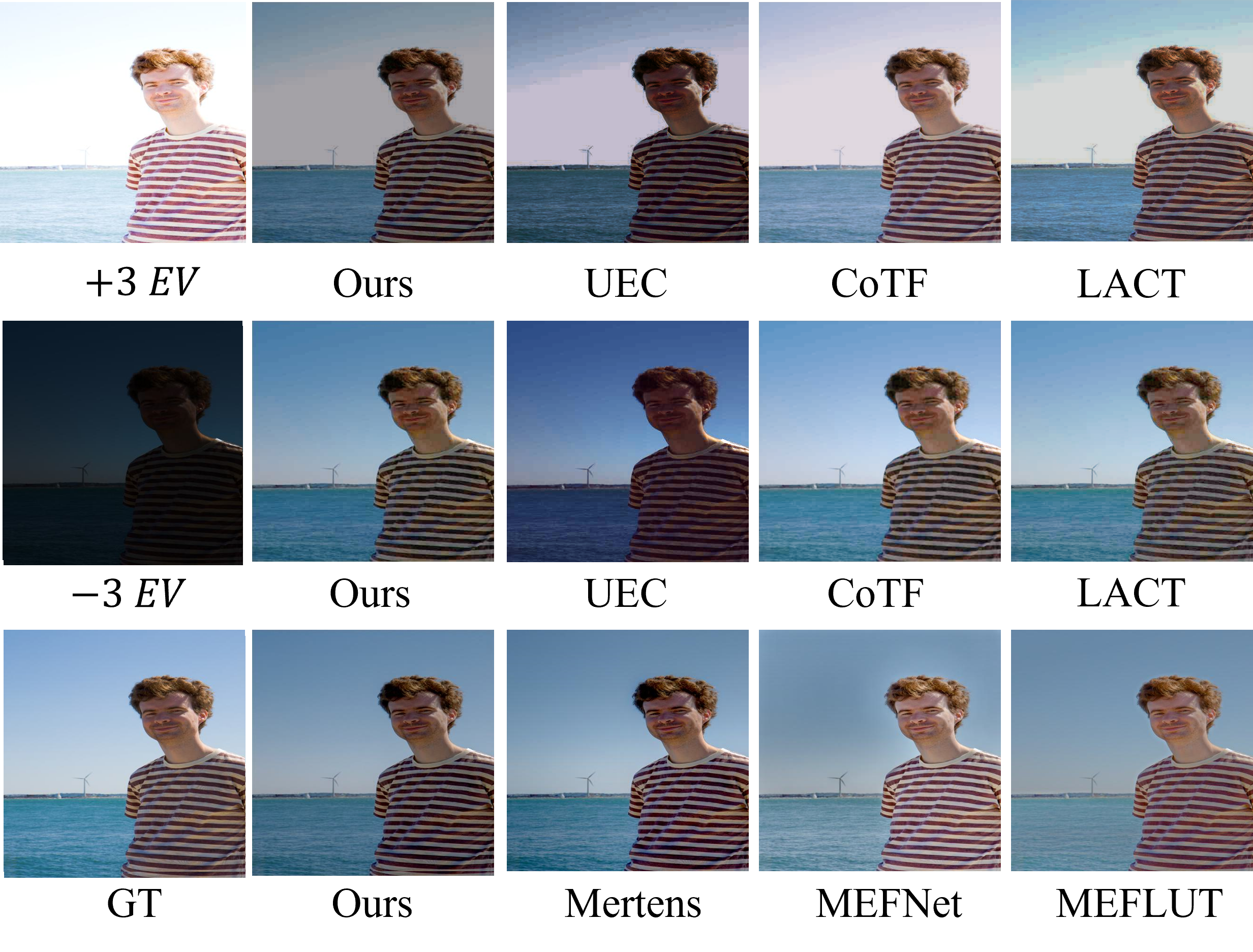}
	\caption{
		 Visual comparison on the Radiometry512 dataset~\cite{uec}. The first row displays the input $+3$ EV image and its corresponding corrected results by different methods, while the second row shows the $-3$ EV image and its corrected versions. The third row presents the ground truth (GT) and the fused results generated by different MEF algorithms.} 
	\label{radio}
\end{figure}

\begin{figure}[t]
	\centering
	\includegraphics[width = \linewidth]{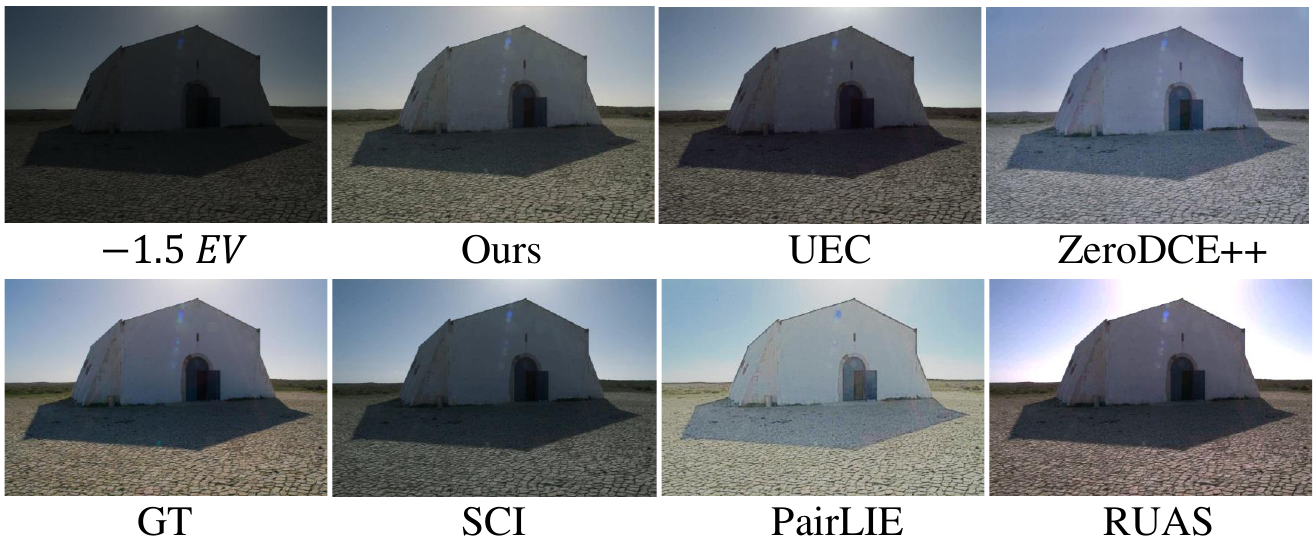}
	\caption{Visual comparison with unsupervised exposure correction and low-light image enhancement methods on MSEC dataset \cite{msec}. } 
	\label{viscomu}
\end{figure}

\begin{figure*}[t]
	\centering
	\includegraphics[width = \linewidth]{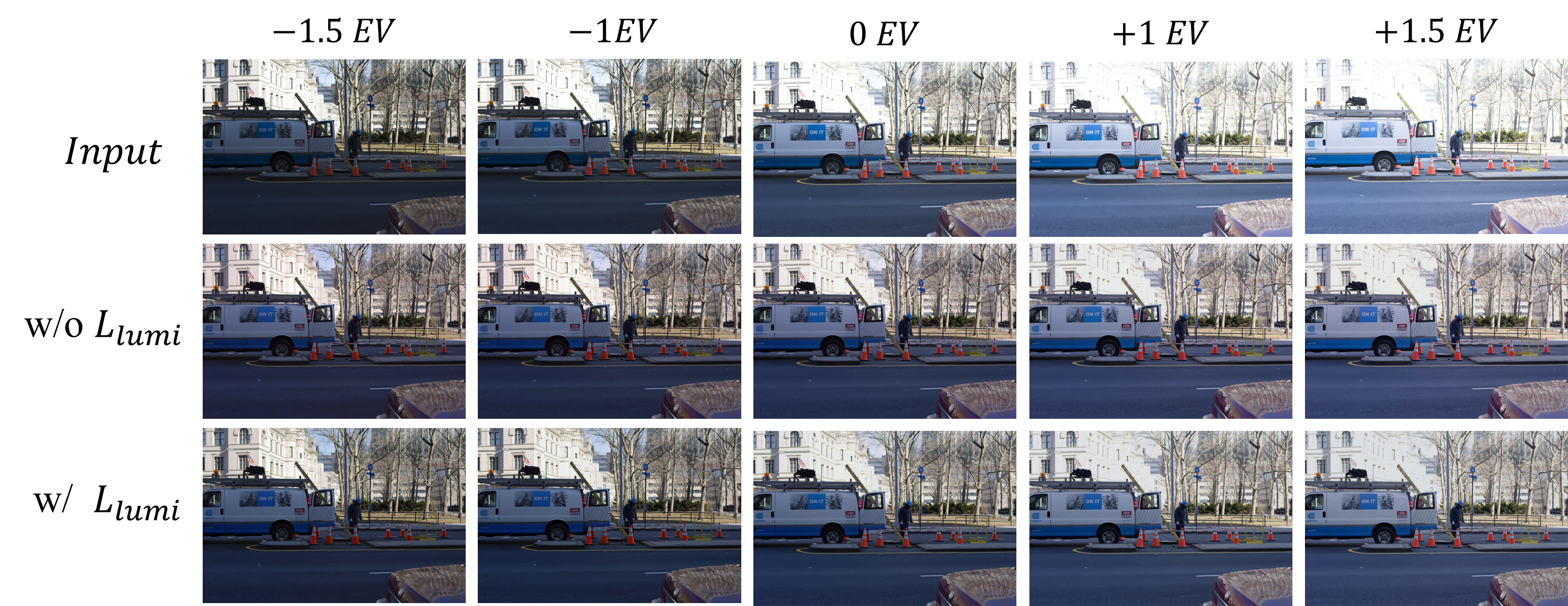}
	\caption{Visual comparison illustrating the impact of the proposed Luminance Ranking Loss.} 
	\label{ablacom}
\end{figure*}

\subsubsection{Multi-Exposure Fusion Results}

We further compare LoopExpose with SOTA MEF methods, including Mertens \cite{mertens}, MEFNet \cite{mefnet}, and MEFLUT \cite{meflut}. Recent MEF algorithms mainly focus on fusing one underexposed and one overexposed image, making them less suitable for rich exposure sequences. We retrain MEFNet and MEFLUT on our reorganized SeqMSEC and SeqRadio datasets for fair comparison.

As presented in Table \ref{mef}, our framework consistently outperforms existing MEF methods significantly across both datasets. These results demonstrate that LoopExpose serves not only as a powerful exposure correction tool, but also as a competitive fusion solution. Its enhanced traditional pipeline, though simple in design, surpasses deep learning-based MEF methods, demonstrating that refining the input before fusion is a highly effective strategy. 

Note that while Mertens performs well in static scenes, it may suffer from ghosting in dynamic scenarios, so these results are primarily for validating the effectiveness of our correction-before-fusion strategy rather than for replacing dedicated MEF pipelines.

\subsubsection{Comparison with Arbitrary-Length Exposure Correction Method}

Table~\ref{tab:sec-mef} reports the quantitative results comparing our proposed LoopExpose with FCNet~\cite{fcnet}, the only existing method capable of arbitrary-length exposure correction. While FCNet is supervised, our method is fully unsupervised. Despite this, LoopExpose achieves superior performance in both SEC and MEF tasks on the MSEC dataset~\cite{msec}, demonstrating that unsupervised frameworks can not only match but even surpass supervised counterparts when designed properly. This further underscores the robustness and generalization ability of LoopExpose under flexible exposure configurations.

\begin{table}[t]
	\centering
	\caption{Impact of the proposed Luminance Ranking Loss on model performance.}
	\label{tab:ablation}

	\footnotesize
	\begin{tabular}{c c c cccc}  
		\toprule[1.5pt]
		\multirow{2}{*}{Setting} 
		& \multirow{2}{*}{$L_{sup}$} 
		& \multirow{2}{*}{$L_{lumi}$} 
		& \multicolumn{2}{c}{SEC} 
		& \multicolumn{2}{c}{MEF} \\
		\cmidrule(lr){4-5} \cmidrule(lr){6-7}
		& & & PSNR & SSIM & PSNR & SSIM \\
		\midrule
		1 & $\checkmark$ &  & 20.7198 & 0.8191 & 21.1146 & 0.8434 \\
		2 & $\checkmark$ & $\checkmark$ &  \textbf{21.0070} & \textbf{0.8257} & \textbf{21.3207} & \textbf{0.8469} \\
		
		\bottomrule[1.5pt]
	\end{tabular}

\end{table}

\subsubsection{Qualitative Comparison}

Figure~\ref{viscom} presents a qualitative comparison of SEC and MEF results on a representative scene from the MSEC dataset~\cite{msec}. As shown, UEC and CoTF both experienced color cast issues at +1.5EV. In contrast, our proposed LoopExpose and LACT remain a balanced color. However, LACT exhibits color noise in some white areas. Moreover, in the fusion results, our method outperforms by preserving both texture and exposure consistency across the entire scene. 
Additionally, Figure~\ref{radio} provides a qualitative comparison on the Radiometry512 dataset~\cite{uec}.  Our method effectively balances the foreground and background, preserving facial texture while maintaining background clarity. This example further demonstrates our model’s ability to generalize across different datasets and exposure conditions.

We further evaluate our approach against a broader range of unsupervised methods. Since most recent unsupervised exposure correction techniques focus primarily on low-light image enhancement (LLIE), we select a dark input image from the MSEC dataset~\cite{msec} to conduct a fair comparison. Figure~\ref{viscomu} provides a qualitative comparison between our method and several unsupervised LLIE techniques~\cite{zerodce++, sci, pairlie, ruas}. For completeness, we also include UEC~\cite{uec} in this comparison. As shown in Figure~\ref{viscomu}, existing LLIE methods often suffer from over-saturation, loss of structural details, or color distortions. In contrast, our method recover more faithful textures and natural colors, effectively balancing brightness and contrast across the scene. These results demonstrate that, although not originally designed for low-light enhancement, our unsupervised exposure correction framework generalizes well to low-light image enhancement.

\subsection{Ablation Study}

\subsubsection{Impact of Luminance Ranking Loss}

We first analyze the contributions of Luminance Ranking Loss, as summarized in Table~\ref{tab:ablation}. Specifically, we evaluate the model using supervised loss $L_{sup}$ alone (Setting 1) versus the combination of supervised loss with the Luminance Ranking Loss $L_{lumi}$ (Setting 2).

Setting 1, utilizing only supervised loss ($L_{sup}$), establishes a solid baseline, demonstrating the effectiveness of fusion-based pseudo-labels for supervision. Setting 2 introduces the additional Luminance Ranking Loss ($L_{lumi}$). The inclusion of $L_{lumi}$ consistently enhances performance in both SEC and MEF tasks. Visual comparisons in Fig.~\ref{ablacom} further illustrate that incorporating luminance constraints helps preserve exposure consistency and improves overall image enhancement quality.

\begin{table}[!t]
	\centering
	\caption{Effectiveness of the multi-stage nested optimization strategy. Stage 1 corresponds to the fixed pseudo-label training (Warm-Up phase), Stage 2 denotes the dynamic pseudo-label refinement (Joint Optimization phase), and Stage 1 + 2 represents the full two-stage nested optimization. The complete scheme achieves the best performance across both SEC and MEF tasks.}
	\label{tab:stage-ablation}

	\footnotesize
	\begin{tabular}{c c c cccc}  
		\toprule[1.5pt]
		\multirow{2}{*}{Setting} 
		& \multirow{2}{*}{Stage 1} 
		& \multirow{2}{*}{Stage 2} 
		& \multicolumn{2}{c}{SEC} 
		& \multicolumn{2}{c}{MEF} \\
		\cmidrule(lr){4-5} \cmidrule(lr){6-7}
		& & & PSNR & SSIM & PSNR & SSIM \\
		\midrule
		1 & $\checkmark$ &              & 20.3615 & 0.8167 & 20.8338 & 0.8398 \\
		2 &              & $\checkmark$ & 20.8338 & 0.8238 & 21.2088 & 0.8455 \\
		3 & $\checkmark$ & $\checkmark$ & \textbf{21.0070} & \textbf{0.8257} & \textbf{21.3207} & \textbf{0.8469}  \\
		\bottomrule[1.5pt]
	\end{tabular}
\end{table}

\subsubsection{Effectiveness of Nested Optimization Strategy}

We further investigate the multi-stage training scheme, highlighting the benefits of the nested optimization approach introduced in this study. The experimental results summarized in Table~\ref{tab:stage-ablation} clearly demonstrate the incremental benefit provided by each stage.

Stage 1 alone (fixed pseudo-label guidance) establishes a reasonable initial performance but is inherently constrained due to static pseudo-labeling. Stage 2, leveraging dynamic pseudo-label updating, significantly improves adaptive capability of the model, thus achieving higher performance. Crucially, the complete two-stage training approach, enabled by our nested optimization framework, yields optimal performance. This underscores the necessity and effectiveness of combining initial stable guidance with subsequent dynamic refinement, ultimately promoting training stability and enhancing final image reconstruction quality.

\section{Conclusion}

In this work, we introduce LoopExpose, a novel unsupervised framework designed to perform reliable exposure correction under diverse lighting conditions. Our nested optimization strategy allows the exposure correction model to be iteratively improved with updated pseudo-labels from a multi-exposure fusion process. This self-reinforcing cycle effectively mitigates the reliance on manually labeled datasets and static fusion baselines. Through comprehensive evaluations on benchmark datasets (MSEC and Radiometry512), LoopExpose consistently outperforms existing unsupervised methods and demonstrates competitive results compared to recent supervised approaches. As a flexible framework, both the SEC and MEF components can be replaced with alternative architectures. In future work, we plan to further improve the training stability. Building upon this, we also aim to explore replacing the fixed MEF module with a learnable fusion model to push the performance ceiling even further.

\bibliographystyle{IEEEtran}
\bibliography{aaai2026.bib}

@article{clahe,
  title={Contrast limited adaptive histogram equalization.},
  author={Zuiderveld, Karel J and others},
  journal={Graphics gems},
  volume={4},
  number={1},
  pages={474--485},
  year={1994},
  publisher={Academic Press, Boston, MA, USA}
}

@article{adaptivehe,
  title={Adaptive histogram equalization and its variations},
  author={Pizer, Stephen M and Amburn, E Philip and Austin, John D and Cromartie, Robert and Geselowitz, Ari and Greer, Trey and ter Haar Romeny, Bart and Zimmerman, John B and Zuiderveld, Karel},
  journal={Computer vision, graphics, and image processing},
  volume={39},
  number={3},
  pages={355--368},
  year={1987},
  publisher={Elsevier}
}

@inproceedings{autoec,
  title={Automatic exposure correction of consumer photographs},
  author={Yuan, Lu and Sun, Jian},
  booktitle={Computer Vision--ECCV 2012: 12th European Conference on Computer Vision, Florence, Italy, October 7-13, 2012, Proceedings, Part IV 12},
  pages={771--785},
  year={2012},
  organization={Springer}
}

@article{naturalnessretinex,
  title={Naturalness preserved enhancement algorithm for non-uniform illumination images},
  author={Wang, Shuhang and Zheng, Jin and Hu, Hai-Miao and Li, Bo},
  journal={IEEE transactions on image processing},
  volume={22},
  number={9},
  pages={3538--3548},
  year={2013},
  publisher={IEEE}
}

@inproceedings{zerodce,
  title={Zero-reference deep curve estimation for low-light image enhancement},
  author={Guo, Chunle and Li, Chongyi and Guo, Jichang and Loy, Chen Change and Hou, Junhui and Kwong, Sam and Cong, Runmin},
  booktitle={Proceedings of the IEEE/CVF conference on computer vision and pattern recognition},
  pages={1780--1789},
  year={2020}
}

@article{zerodce++,
  title={Learning to enhance low-light image via zero-reference deep curve estimation},
  author={Li, Chongyi and Guo, Chunle and Loy, Chen Change},
  journal={IEEE transactions on pattern analysis and machine intelligence},
  volume={44},
  number={8},
  pages={4225--4238},
  year={2021},
  publisher={IEEE}
}

@InProceedings{noise2noise,
  title = 	 {{N}oise2{N}oise: Learning Image Restoration without Clean Data},
  author =       {Lehtinen, Jaakko and Munkberg, Jacob and Hasselgren, Jon and Laine, Samuli and Karras, Tero and Aittala, Miika and Aila, Timo},
  booktitle = 	 {Proceedings of the 35th International Conference on Machine Learning},
  pages = 	 {2965--2974},
  year = 	 {2018},
  editor = 	 {Dy, Jennifer and Krause, Andreas},
  volume = 	 {80},
  series = 	 {Proceedings of Machine Learning Research},
  month = 	 {10--15 Jul},
  publisher =    {PMLR}
}

@inproceedings{mertens,
  title={Exposure fusion: A simple and practical alternative to high dynamic range photography},
  author={Mertens, Tom and Kautz, Jan and Van Reeth, Frank},
  booktitle={Computer graphics forum},
  volume={28},
  number={1},
  pages={161--171},
  year={2009},
  organization={Wiley Online Library}
}

@article{gf,
  title={Image fusion with guided filtering},
  author={Li, Shutao and Kang, Xudong and Hu, Jianwen},
  journal={IEEE Transactions on Image processing},
  volume={22},
  number={7},
  pages={2864--2875},
  year={2013},
  publisher={IEEE}
}

@inproceedings{msec,
  title={Learning multi-scale photo exposure correction},
  author={Afifi, Mahmoud and Derpanis, Konstantinos G and Ommer, Bjorn and Brown, Michael S},
  booktitle={Proceedings of the IEEE/CVF Conference on Computer Vision and Pattern Recognition},
  pages={9157--9167},
  year={2021}
}

@inproceedings{CMEC,
  title={Learning Exposure Correction Via Consistency Modeling.},
  author={Nsampi, Ntumba Elie and Hu, Zhongyun and Wang, Qing},
  booktitle={BMVC},
  pages={12},
  year={2021}
}

@inproceedings{enc,
  title={Exposure normalization and compensation for multiple-exposure correction},
  author={Huang, Jie and Liu, Yajing and Fu, Xueyang and Zhou, Man and Wang, Yang and Zhao, Feng and Xiong, Zhiwei},
  booktitle={Proceedings of the IEEE/CVF Conference on Computer Vision and Pattern Recognition},
  pages={6043--6052},
  year={2022}
}

@inproceedings{ecl,
  title={Exposure-consistency representation learning for exposure correction},
  author={Huang, Jie and Zhou, Man and Liu, Yajing and Yao, Mingde and Zhao, Feng and Xiong, Zhiwei},
  booktitle={Proceedings of the 30th ACM International Conference on Multimedia},
  pages={6309--6317},
  year={2022}
}

@inproceedings{fecnet,
  title={Deep fourier-based exposure correction network with spatial-frequency interaction},
  author={Huang, Jie and Liu, Yajing and Zhao, Feng and Yan, Keyu and Zhang, Jinghao and Huang, Yukun and Zhou, Man and Xiong, Zhiwei},
  booktitle={European Conference on Computer Vision},
  pages={163--180},
  year={2022},
  organization={Springer}
}

@inproceedings{lcdp,
  title={Local color distributions prior for image enhancement},
  author={Wang, Haoyuan and Xu, Ke and Lau, Rynson WH},
  booktitle={European conference on computer vision},
  pages={343--359},
  year={2022},
  organization={Springer}
}

@inproceedings{da,
  title={Decoupling-and-aggregating for image exposure correction},
  author={Wang, Yang and Peng, Long and Li, Liang and Cao, Yang and Zha, Zheng-Jun},
  booktitle={Proceedings of the IEEE/CVF conference on computer vision and pattern recognition},
  pages={18115--18124},
  year={2023}
}

@inproceedings{erl,
  title={Learning sample relationship for exposure correction},
  author={Huang, Jie and Zhao, Feng and Zhou, Man and Xiao, Jie and Zheng, Naishan and Zheng, Kaiwen and Xiong, Zhiwei},
  booktitle={Proceedings of the IEEE/CVF conference on computer vision and pattern recognition},
  pages={9904--9913},
  year={2023}
}

@inproceedings{lact,
  title={Luminance-aware color transform for multiple exposure correction},
  author={Baek, Jong-Hyeon and Kim, DaeHyun and Choi, Su-Min and Lee, Hyo-jun and Kim, Hanul and Koh, Yeong Jun},
  booktitle={Proceedings of the IEEE/CVF international conference on computer vision},
  pages={6156--6165},
  year={2023}
}

@inproceedings{mmht,
  title={Fearless luminance adaptation: A macro-micro-hierarchical transformer for exposure correction},
  author={Li, Gehui and Liu, Jinyuan and Ma, Long and Jiang, Zhiying and Fan, Xin and Liu, Risheng},
  booktitle={Proceedings of the 31st ACM International Conference on Multimedia},
  pages={7304--7313},
  year={2023}
}

@inproceedings{cotf,
  title={Real-time exposure correction via collaborative transformations and adaptive sampling},
  author={Li, Ziwen and Zhang, Feng and Cao, Meng and Zhang, Jinpu and Shao, Yuanjie and Wang, Yuehuan and Sang, Nong},
  booktitle={Proceedings of the IEEE/CVF Conference on Computer Vision and Pattern Recognition},
  pages={2984--2994},
  year={2024}
}

@inproceedings{uec,
  title={Unsupervised Exposure Correction},
  author={Cui, Ruodai and Niu, Li and Hu, Guosheng},
  booktitle={European Conference on Computer Vision},
  pages={252--268},
  year={2024},
  organization={Springer}
}

@inproceedings{dual,
  title={Dual illumination estimation for robust exposure correction},
  author={Zhang, Qing and Nie, Yongwei and Zheng, Wei-Shi},
  booktitle={Computer graphics forum},
  volume={38},
  number={7},
  pages={243--252},
  year={2019},
  organization={Wiley Online Library}
}

@article{opencv,
    author = {Bradski, G.},
    journal = {Dr. Dobb's Journal of Software Tools},
    title = {{The OpenCV Library}},
    year = {2000}
}

@InProceedings{csec,
    author    = {Li, Yiyu and Xu, Ke and Hancke, Gerhard Petrus and Lau, Rynson W.H.},
    title     = {Color Shift Estimation-and-Correction for Image Enhancement},
    booktitle = {Proceedings of the IEEE/CVF Conference on Computer Vision and Pattern Recognition (CVPR)},
    month     = {June},
    year      = {2024},
    pages     = {25389-25398}
}

@inproceedings{cyclegan,
  title={Unpaired image-to-image translation using cycle-consistent adversarial networks},
  author={Zhu, Jun-Yan and Park, Taesung and Isola, Phillip and Efros, Alexei A},
  booktitle={Proceedings of the IEEE international conference on computer vision},
  pages={2223--2232},
  year={2017}
}

@inproceedings{hdrmodel,
  title={The state of the art in HDR deghosting: A survey and evaluation},
  author={Tursun, Okan Tarhan and Aky{\"u}z, Ahmet O{\u{g}}uz and Erdem, Aykut and Erdem, Erkut},
  booktitle={Computer Graphics Forum},
  volume={34},
  number={2},
  pages={683--707},
  year={2015},
  organization={Wiley Online Library}
}

@article{modelingResponse,
  title={Modeling the space of camera response functions},
  author={Grossberg, Michael D and Nayar, Shree K},
  journal={IEEE transactions on pattern analysis and machine intelligence},
  volume={26},
  number={10},
  pages={1272--1282},
  year={2004},
  publisher={IEEE}
}

@inproceedings{debevec,
author = {Debevec, Paul E. and Malik, Jitendra},
title = {Recovering high dynamic range radiance maps from photographs},
year = {1997},
isbn = {0897918967},
publisher = {ACM Press/Addison-Wesley Publishing Co.},
address = {USA},
booktitle = {Proceedings of the 24th Annual Conference on Computer Graphics and Interactive Techniques},
pages = {369–378},
numpages = {10},
series = {SIGGRAPH '97}
}

@inproceedings{fivek,
  title={Learning photographic global tonal adjustment with a database of input/output image pairs},
  author={Bychkovsky, Vladimir and Paris, Sylvain and Chan, Eric and Durand, Fr{\'e}do},
  booktitle={CVPR 2011},
  pages={97--104},
  year={2011},
  organization={IEEE}
}

@misc{fcnet,
      title={FCNet: A Convolutional Neural Network for Arbitrary-Length Exposure Estimation}, 
      author={Jin Liang and Yuchen Yang and Anran Zhang and Jun Xu and Hui Li and Xiantong Zhen},
      year={2023},
      eprint={2203.03624},
      archivePrefix={arXiv},
      primaryClass={eess.IV},
      url={https://arxiv.org/abs/2203.03624}, 
}

@article{traditionalBL,
  title={Mathematical programs with optimization problems in the constraints},
  author={Bracken, Jerome and McGill, James T},
  journal={Operations research},
  volume={21},
  number={1},
  pages={37--44},
  year={1973},
  publisher={INFORMS}
}

@inproceedings{2021bilevel,
  title={Bilevel optimization: Convergence analysis and enhanced design},
  author={Ji, Kaiyi and Yang, Junjie and Liang, Yingbin},
  booktitle={International conference on machine learning},
  pages={4882--4892},
  year={2021},
  organization={PMLR}
}

@inproceedings{stabilityBL,
  title={Fine-grained analysis of stability and generalization for stochastic bilevel Optimization},
  author={Zhang, Xuelin and Chen, Hong and Gu, Bin and Gong, Tieliang and Zheng, Feng},
  booktitle={Proceedings of the Thirty-Third International Joint Conference on Artificial Intelligence},
  pages={5508--5516},
  year={2024}
}

@article{mefgan,
  title={MEF-GAN: Multi-exposure image fusion via generative adversarial networks},
  author={Xu, Han and Ma, Jiayi and Zhang, Xiao-Ping},
  journal={IEEE Transactions on Image Processing},
  volume={29},
  pages={7203--7216},
  year={2020},
  publisher={IEEE}
}

@inproceedings{meflut,
  title={Meflut: Unsupervised 1d lookup tables for multi-exposure image fusion},
  author={Jiang, Ting and Wang, Chuan and Li, Xinpeng and Li, Ru and Fan, Haoqiang and Liu, Shuaicheng},
  booktitle={Proceedings of the IEEE/CVF International Conference on Computer Vision},
  pages={10542--10551},
  year={2023}
}

@article{mefnet,
  title={Deep guided learning for fast multi-exposure image fusion},
  author={Ma, Kede and Duanmu, Zhengfang and Zhu, Hanwei and Fang, Yuming and Wang, Zhou},
  journal={IEEE Transactions on Image Processing},
  volume={29},
  pages={2808--2819},
  year={2019},
  publisher={IEEE}
}

@inproceedings{deepfuse,
  title={Deepfuse: A deep unsupervised approach for exposure fusion with extreme exposure image pairs},
  author={Ram Prabhakar, K and Sai Srikar, V and Venkatesh Babu, R},
  booktitle={Proceedings of the IEEE international conference on computer vision},
  pages={4714--4722},
  year={2017}
}

@inproceedings{transmef,
  title={Transmef: A transformer-based multi-exposure image fusion framework using self-supervised multi-task learning},
  author={Qu, Linhao and Liu, Shaolei and Wang, Manning and Song, Zhijian},
  booktitle={Proceedings of the AAAI conference on artificial intelligence},
  volume={36},
  number={2},
  pages={2126--2134},
  year={2022}
}

@article{ifcnn,
  title={IFCNN: A general image fusion framework based on convolutional neural network},
  author={Zhang, Yu and Liu, Yu and Sun, Peng and Yan, Han and Zhao, Xiaolin and Zhang, Li},
  journal={Information Fusion},
  volume={54},
  pages={99--118},
  year={2020},
  publisher={Elsevier}
}

@article{pairwisecomparison,
  title={A scaling method for priorities in hierarchical structures},
  author={Saaty, Thomas L},
  journal={Journal of mathematical psychology},
  volume={15},
  number={3},
  pages={234--281},
  year={1977},
  publisher={Elsevier}
}

@article{dncnn,
  title={Beyond a gaussian denoiser: Residual learning of deep cnn for image denoising},
  author={Zhang, Kai and Zuo, Wangmeng and Chen, Yunjin and Meng, Deyu and Zhang, Lei},
  journal={IEEE Transactions on Image Processing},
  volume={26},
  number={7},
  pages={3142--3155},
  year={2017},
  publisher={IEEE}
}

@article{3dlut,
  title={Learning image-adaptive 3d lookup tables for high performance photo enhancement in real-time},
  author={Zeng, Hui and Cai, Jianrui and Li, Lida and Cao, Zisheng and Zhang, Lei},
  journal={IEEE Transactions on Pattern Analysis and Machine Intelligence},
  volume={44},
  number={4},
  pages={2058--2073},
  year={2020},
  publisher={IEEE}
}

@inproceedings{wvm,
  title={A weighted variational model for simultaneous reflectance and illumination estimation},
  author={Fu, Xueyang and Zeng, Delu and Huang, Yue and Zhang, Xiao-Ping and Ding, Xinghao},
  booktitle={Proceedings of the IEEE conference on computer vision and pattern recognition},
  pages={2782--2790},
  year={2016}
}

@inproceedings{ruas,
  title={Retinex-inspired unrolling with cooperative prior architecture search for low-light image enhancement},
  author={Liu, Risheng and Ma, Long and Zhang, Jiaao and Fan, Xin and Luo, Zhongxuan},
  booktitle={Proceedings of the IEEE/CVF conference on computer vision and pattern recognition},
  pages={10561--10570},
  year={2021}
}

@inproceedings{sci,
  title={Toward fast, flexible, and robust low-light image enhancement},
  author={Ma, Long and Ma, Tengyu and Liu, Risheng and Fan, Xin and Luo, Zhongxuan},
  booktitle={Proceedings of the IEEE/CVF conference on computer vision and pattern recognition},
  pages={5637--5646},
  year={2022}
}

@inproceedings{pairlie,
  title={Learning a simple low-light image enhancer from paired low-light instances},
  author={Fu, Zhenqi and Yang, Yan and Tu, Xiaotong and Huang, Yue and Ding, Xinghao and Ma, Kai-Kuang},
  booktitle={Proceedings of the IEEE/CVF conference on computer vision and pattern recognition},
  pages={22252--22261},
  year={2023}
}

@inproceedings{noise2void,
  title={Noise2void-learning denoising from single noisy images},
  author={Krull, Alexander and Buchholz, Tim-Oliver and Jug, Florian},
  booktitle={Proceedings of the IEEE/CVF conference on computer vision and pattern recognition},
  pages={2129--2137},
  year={2019}
}

@inproceedings{neighbor2neighbor,
  title={Neighbor2neighbor: Self-supervised denoising from single noisy images},
  author={Huang, Tao and Li, Songjiang and Jia, Xu and Lu, Huchuan and Liu, Jianzhuang},
  booktitle={Proceedings of the IEEE/CVF conference on computer vision and pattern recognition},
  pages={14781--14790},
  year={2021}
}

@book{adobe,
  title={Real World Camera Raw with Adobe Photoshop CS5},
  author={Schewe, Jeff and Fraser, Bruce},
  year={2010},
  publisher={Pearson Education}
}

@article{anisotropicRetinex,
  title={Anisotropic pth-order TV-based Retinex decomposition with adaptive reflectance regularizer for low-light image enhancement},
  author={Hsieh, Po-Wen and Yang, Suh-Yuh},
  journal={Pattern Recognition},
  pages={112468},
  year={2025},
  publisher={Elsevier}
}

@article{przero,
  title={Zero-shot image denoising with hollow pair sampling and noise-aware attention},
  author={Wang, Sheng and Zhao, Chaoyue and Wang, Qiao and Liu, Mingzhi and Mou, Chao and Xu, Fu},
  journal={Pattern Recognition},
  pages={111779},
  year={2025},
  publisher={Elsevier}
}

@article{ctnet,
  title={CTNet: Color transformation network for low-light image enhancement},
  author={Xie, Lidong and Cong, Runmin and Dai, Ju and Yang, Wenhan and Pan, Junjun and Wu, Hao},
  journal={Pattern Recognition},
  pages={112360},
  year={2025},
  publisher={Elsevier}
}

@article{noiseretinex,
  title={Nonuniform low-light image enhancement via noise-aware decomposition and adaptive correction},
  author={Huang, Jiancai and Jiang, Zhaohui and Liu, Xingjian and Tan, Yap-Peng and Gui, Weihua},
  journal={Pattern Recognition},
  pages={112370},
  year={2025},
  publisher={Elsevier}
}

@article{retina,
  title={Retina Adaptation Network for Low-light Image Enhancement},
  author={Zhang, Houwang and Zhang, Xian-Shi and Yang, Kai-Fu and Li, Yong-Jie and Chan, Leanne Lai Hang},
  journal={Pattern Recognition},
  pages={112629},
  year={2025},
  publisher={Elsevier}
}

@article{prdiffretinex,
  title={Retinex-guided generative diffusion prior for low-light image enhancement},
  author={Zhao, Zunjin and Shi, Daming},
  journal={Pattern Recognition},
  pages={112421},
  year={2025},
  publisher={Elsevier}
}

@article{diffdarkretinex,
  title={DiffDark: Multi-prior integration driven diffusion model for low-light image enhancement},
  author={Hu, Renzhi and Luo, Ting and Jiang, Gangyi and Chen, Yeyao and Xu, Haiyong and Liu, Leiming and He, Zhouyan},
  journal={Pattern Recognition},
  pages={111814},
  year={2025},
  publisher={Elsevier}
}

@article{exposurediff,
  title={Exposure difference network for low-light image enhancement},
  author={Jiang, Shengqin and Mei, Yongyue and Wang, Peng and Liu, Qingshan},
  journal={Pattern Recognition},
  volume={156},
  pages={110796},
  year={2024},
  publisher={Elsevier}
}

@article{gamma,
  title={Low-light image enhancement using gamma correction prior in mixed color spaces},
  author={Jeon, Jong Ju and Park, Jun Young and Eom, Il Kyu},
  journal={Pattern Recognition},
  volume={146},
  pages={110001},
  year={2024},
  publisher={Elsevier}
}

@article{prhdr,
  title={HDR reconstruction from a single exposure LDR using texture and structure dual-stream generation},
  author={Chen, Yu-Hsiang and Ruan, Shanq-Jang},
  journal={Pattern Recognition},
  volume={159},
  pages={111127},
  year={2025},
  publisher={Elsevier}
}

\newpage

\begin{IEEEbiography}[{\includegraphics[width=1in,height=1.25in,clip,keepaspectratio]{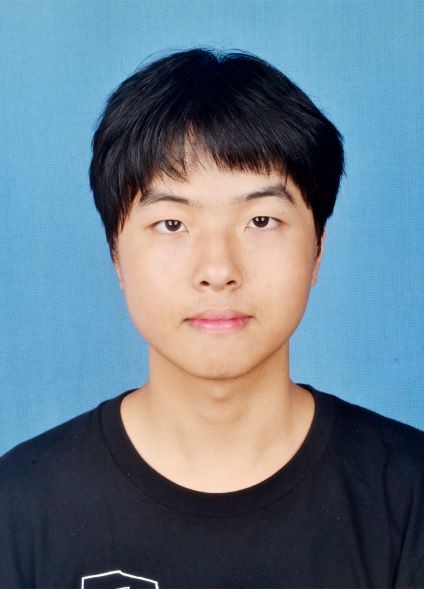}}]{Ao Li} received the B.E. degree in Artificial Intelligence from Xidian University, Xi'an, China, in 2024. He is currently pursuing the Ph.D. degree with the School of Artificial Intelligence, Xidian University, Xi’an, China. His research interests include low-level vision, computer vision and deep learning.
\end{IEEEbiography}

\begin{IEEEbiography}[{\includegraphics[width=1in,height=1.25in,clip,keepaspectratio]{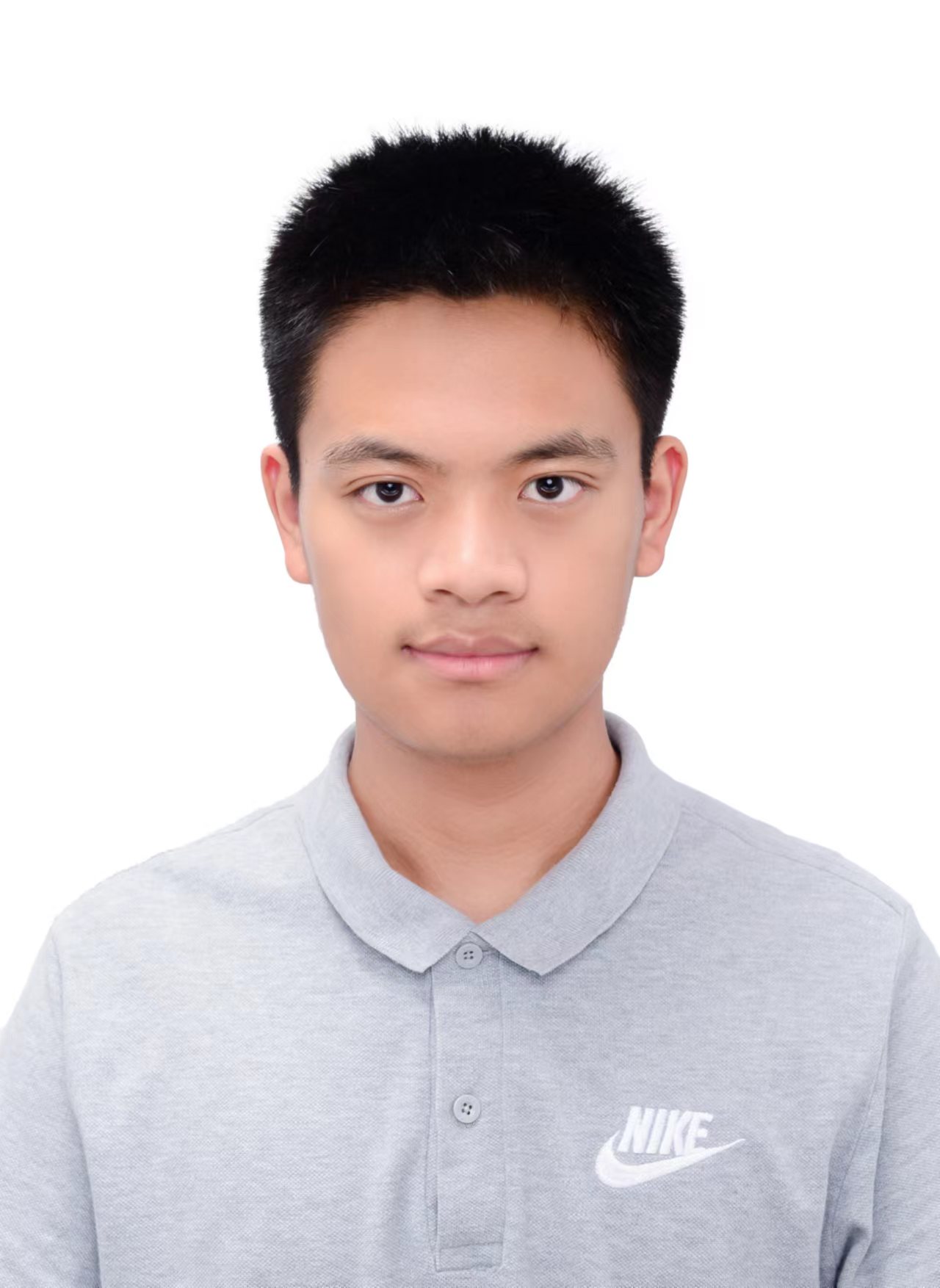}}]{Chen Chen} is currently pursuing the B.S. degree with the School of Information Science and Engineering, Dalian University of Technology, Dalian, China. He is interested in image processing and deep learning.
\end{IEEEbiography}

\begin{IEEEbiography}[{\includegraphics[width=1in,height=1.25in,clip,keepaspectratio]{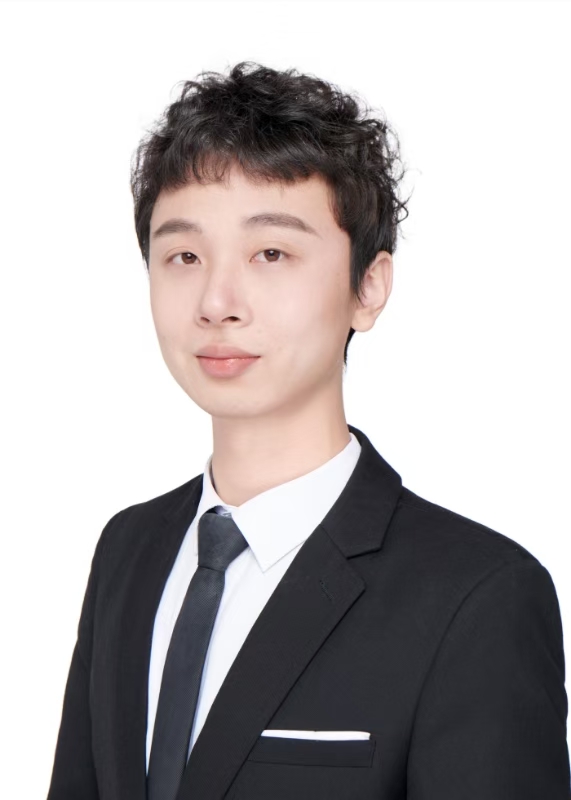}}]{Zhenyu Wang} received the Ph.D. degree from the School of Artificial Intelligence, Xidian University, Xi’an, China, in 2023, where he received the B.S. degree from the School of Microelectronics. He is currently the postdoctor in Hangzhou Institute of Technology, Xidian University. His research interests include deep
learning, computer vision, neural networks, and model compression.
\end{IEEEbiography}

\begin{IEEEbiography}[{\includegraphics[width=1in,height=1.25in,clip,keepaspectratio]{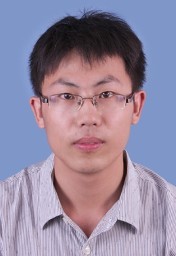}}]{Tao Huang} received his Ph.D. degree from Xidian University in June 2018. From July 2018 to November 2024, he worked at Alibaba Group and Ant Group, engaging in the implementation of computer vision algorithms. Since 2024, he has served as an Associate Research Fellow at the Hangzhou Research Institute of Xidian University, with research interests in computer vision and multimodal content understanding.
\end{IEEEbiography}

\begin{IEEEbiography}[{\includegraphics[width=1in,height=1.25in,clip,keepaspectratio]{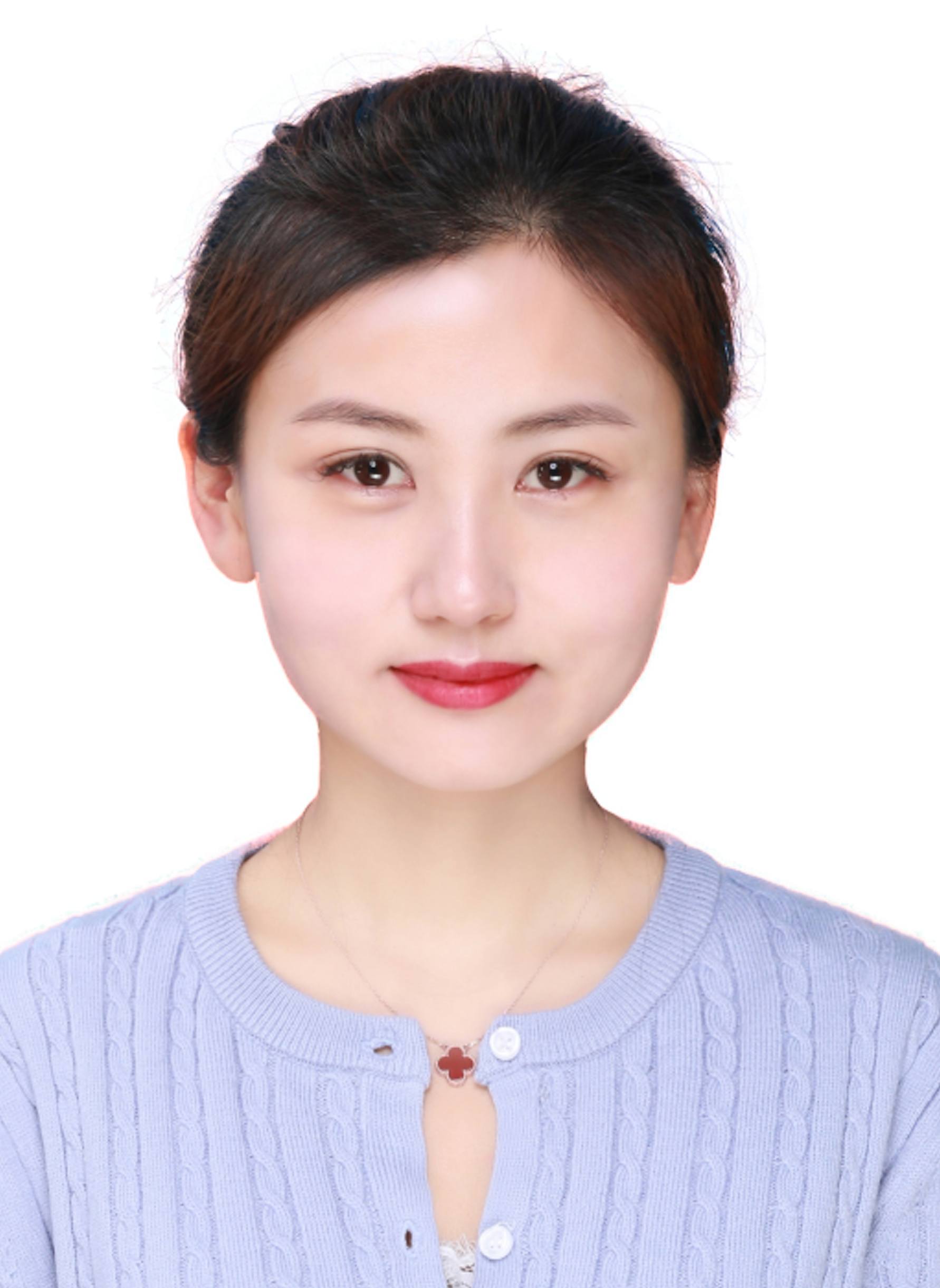}}]{Fangfang Wu}
received the B.S. degree in electrical engineering, the M.S. degree in intelligent information processing, and the Ph.D. degree in electronics science and technology from Xidian University, Xi’an, China, in 2008, 2011, and 2021, respectively. In 2022, she joined the School of Computer Science and Technology, Xidian University. Her research interests include image restoration, hyperspectral imaging and deep image representation.
\end{IEEEbiography}

\begin{IEEEbiography}[{\includegraphics[width=1in,height=1.25in,clip,keepaspectratio]{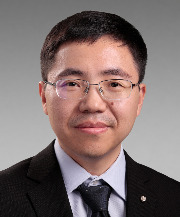}}]{Weisheng Dong}(Member, IEEE)
received the B.S. degree in electronic engineering from the Huazhong University of Science and Technology, Wuhan, China, in 2004, and the Ph.D. degree in circuits and system from Xidian University, Xi'an, China, in 2010. He was a Visiting Student with Microsoft Research Asia, Beijing, China, in 2006. From 2009 to 2010, he was a Research Assistant with the Department of Computing, Hong Kong Polytechnic University, Hong Kong. In 2010, he joined the School of Electronic Engineering, Xidian University, as a Lecturer, where he has been a Professor since 2016. His research interests include inverse problems in image processing, sparse signal representation, and image compression. He was a recipient of the Best Paper Award at the SPIE Visual Communication and Image Processing (VCIP) in 2010.  He has served as an associate editor of IEEE Transactions on Image Processing and is currently an associate editor of SIAM Journal of Imaging Sciences.
\end{IEEEbiography}

\vfill

\end{document}